\documentclass[10pt]{article}
\usepackage{amsfonts}
\usepackage{epsfig}

\usepackage{color}

\begin{document}

\title{Malagasy Dialects and the Peopling of Madagascar}

\author{Maurizio Serva${}^{(1)}$,    
Filippo Petroni${}^{(2)}$, \\
Dima Volchenkov${}^{(3)}$  and 
S{\o}ren Wichmann${}^{(4)}$ 
}
\maketitle
{\scriptsize
\begin{enumerate}
\item[${}^{(1)}$]{\it Dipartimento di Matematica, Universit\`{a} dell'Aquila, 
I-67010 L'Aquila, Italy,}
\item[${}^{(2)}$]{\it Facolt\`a di Economia, Universit\`{a} di Cagliari,
I-09123 Cagliari, Italy, }
\item[${}^{(3)}$]{\it  Center of Excellence Cognitive
Interaction Technology, Universit\"{a}t Bielefeld, Postfach 10 01 31, 33501 Bielefeld, Germany,}
\item[${}^{(4)}$]{\it Max Planck Institute for Evolutionary Anthropology,  
D-04103 Leipzig, Germany. }
\end{enumerate}
}

\begin{flushleft}
\end{flushleft}

\begin{abstract}
The origin of Malagasy DNA is half African and half Indonesian, nevertheless the Malagasy 
language, spoken by the entire population, belongs to the Austronesian family. 
The language most closely related to Malagasy is Maanyan (Greater Barito East group of 
the Austronesian family), but related languages are also in Sulawesi, Malaysia and Sumatra.
For this reason, and because Maanyan is spoken by a population which lives along the 
Barito river in Kalimantan and which does not possess the necessary skill for long maritime 
navigation, the ethnic composition of the Indonesian colonizers is still unclear. 

There is a general consensus that Indonesian sailors reached Madagascar by a 
maritime trek, but the time, the path and the landing area of the first colonization are 
all disputed. 
In this research we try to answer these problems together with other ones, such as
the historical configuration of Malagasy dialects, by types of analysis related to 
lexicostatistics and glottochronology which draw upon the automated method recently 
proposed by the authors \cite{Serva:2008, Holman:2008, Petroni:2008, Bakker:2009}.
The data were collected by the first author at the beginning of 2010 with the invaluable help 
of  Joselin\`a Soafara N\'er\'e and consist of Swadesh lists of 200 items for 23 dialects 
covering all areas of the Island.

\end{abstract}

\section{\label{sec:Introduction}Introduction}
\noindent
The genetic make-up of Malagasy people exhibits almost equal proportions of African and 
Indonesian heritage \cite{Hurles:2005}. Nevertheless, as was suggested already in 
\cite{Houtman:1603}, Malagasy and its dialects have relatives among languages belonging to 
what is today known as the Austronesian linguistic family. This was firmly established in 
\cite{Tuuk:1864}, and \cite{Dahl:1951} pointed out a particularly close relationship 
between Malagasy and Maanyan of south-eastern Kalimantan, which share about 45\% their 
basic vocabulary \cite{Dyen:1953}. But Malagasy also bears similarities to languages in 
Sulawesi, Malaysia and Sumatra, including loanwords from Malay, Javanese, and one (or 
more) language(s) of south Sulawesi \cite{Adelaar:2009}. 
Furthermore, it contains an African component in the vocabulary, especially as regards 
faunal names \cite{Blench and Walsh:2009}.
For this reason, the history of Madagascar peopling and settlement 
is subject to alternative interpretations among scholars. 
It seems that Indonesian sailors reached Madagascar by a maritime trek at a time some 
one to two thousand years ago (the exact time is subject to debate), but it is not clear 
whether there were multiple settlements or just a single one. 
Additional questions are raised by the fact that the Maanyan speakers live along the 
rivers of Kalimantan and have not in historical times possessed the necessary skills for 
long-distance maritime navigation. A possible explanation is that the ancestors of the 
Malagasy did not themselves navigate the boat(s) that took them to Madagascar, but were 
brought as subordinates of Malay sailors \cite{Adelaar:1995b}. If this is the case, then 
Malagasy dialects are expected to show influence from Malay in addition to having a 
component similar to Maanyan. 
While the origin of Malagasy is thus not completely clarified there are also doubts 
relating to the arrival scenario. 
Some scholars \cite{Adelaar:2009} consider it most likely that 
the settlement of the island took place only after an initial arrival on the African 
mainland, while others assume that the island was settled directly, without this detour. 
Finally, to date no satisfactory internal classification of the Malagasy dialects has been 
proposed. To summarize, it would be desirable to know more about (1) when the migration to 
Madagascar took place, (2) how Malagasy is related to other Austronesian languages, (3) 
the historical configuration of Malagasy dialects, and (4) where the original settlement 
of the Malagasy people took place. 

Our research addresses these four problems through the application of new quantitative 
methodologies inspired by, but nevertheless different from, classical lexicostatistics 
and glottochronology \cite{Serva:2008, Holman:2008, Petroni:2008, Bakker:2009}.

The data, collected  during the beginning of 2010, consist of 200-item 
Swadesh word lists for 23 dialects of Malagasy from all areas of the island. 
A practical orthography
which corresponds to the orthographical conventions of standard Malagasy has been used.
Most of the informants were able to write the words directly 
using these conventions, while a few of them benefited from the help of one ore more 
fellow townsmen. 
A cross-checking of each dialect list was done by eliciting data 
separately from two different consultants.
Details about the speakers who furnished the data are provided in Appendix D.
This dataset probably represents the largest collection available of comparative Swadesh 
lists for Malagasy (see Fig.~\ref{Fig_3} for the locations). 
The lists can be found in the database \cite{Serva:2011}.

While there are linguistic as well as geographical and temporal dimensions to the issues 
addressed in this paper, all strands of the investigation are rooted in an automated 
comparison of words through a specific version of the so-called Levenshtein or 'edit' 
distance (henceforth LD) \cite{Levenshtein:1966}. The version we use here was introduced
by \cite{Serva:2008, Petroni:2008} and consists of the following 
procedure. Words referring to the same concept for a given pair of dialects are compared 
with a view to how easily the word in dialect A is transformed into the corresponding word 
in dialect B. Steps allowed in the transformations are: insertions, deletions, and 
substitutions. The LD is then calculated as the minimal number of such steps required 
to completely transform one word into the other. 
Calculating the distance measure that we use (the 'normalized Levenshtein distance', or LDN),
requires one more operation: the 'raw LD' is 
divided by the length (in terms of segments) of the longer of the two words compared. 
This operation produces LDN values between 0\% and 100\%, and takes into account 
variable word lengths: if one or both of the words compared happen to be relatively long, 
the LD is prone to be higher than if they both happen to be short, so without the 
normalization the distance values would not be comparable. Finally we average the LDN's for 
all 200 pairs of words compared to 
obtain a distance value characterizing the overall difference between a pair of dialects 
(see Appendix A for a compact mathematical definition and a table with all distances.).

Thus, the Levenshtein distance is sensitive to both lexical replacement and phonological 
change and therefore differs from the cognate counting 
procedure of classical lexicostatistics even if the results are usually roughly equivalent. 

The first use of the pairwise distances is to derive a classification of the dialects.
For this purpose we adopt a multiple strategy in order to extract a maximum of information
from the set of pairwise distances. 
We first obtain a tree representation of the set by using two different standard 
phylogenetic algorithms,
then we adopt a strategy (SCA) which, analogously to a principal components approach,
represents the set in terms of geometrical relations. 
The SCA analysis also provides the tool for a dating of the landing of Malagasy ancestors
on the island.
The landing area is established assuming that a linguistic homeland is the area
exhibiting the maximum of current linguistic diversity. Diversity is measured by
comparing lexical and geographical distances. Finally, we perform a comparison 
of all variants with some other Austronesian languages, in particular with Malay and Maanyan.

For the purpose of the external comparison of Malagasy variants 
with other Austronesian languages we draw upon {\it The Austronesian Basic Vocabulary 
Database} \cite{Greenhill:2009}. Since the wordlists in this database do not always 
contain all the 200 items of our (and Swadesh) lists they are supplemented by various 
sources, including database of the Automated Similarity Judgment Program (ASJP) 
\cite{Wichmann:2010c}.

\section{\label{sec:The_internal_classification_of_Malagasy}
The internal classification of Malagasy}
\subsection{\label{subsec:Our_results}Our results}
\noindent
In this section we present two different classificatory trees for the 23 Malagasy dialects 
obtained through applying two different phylogenetic algorithms to the set of pairwise 
distances resulting from comparing our 200-item word lists through the normalized 
Levenshtein distance (LDN).

\begin{figure}[ht]
\begin{center}
\epsfig{file=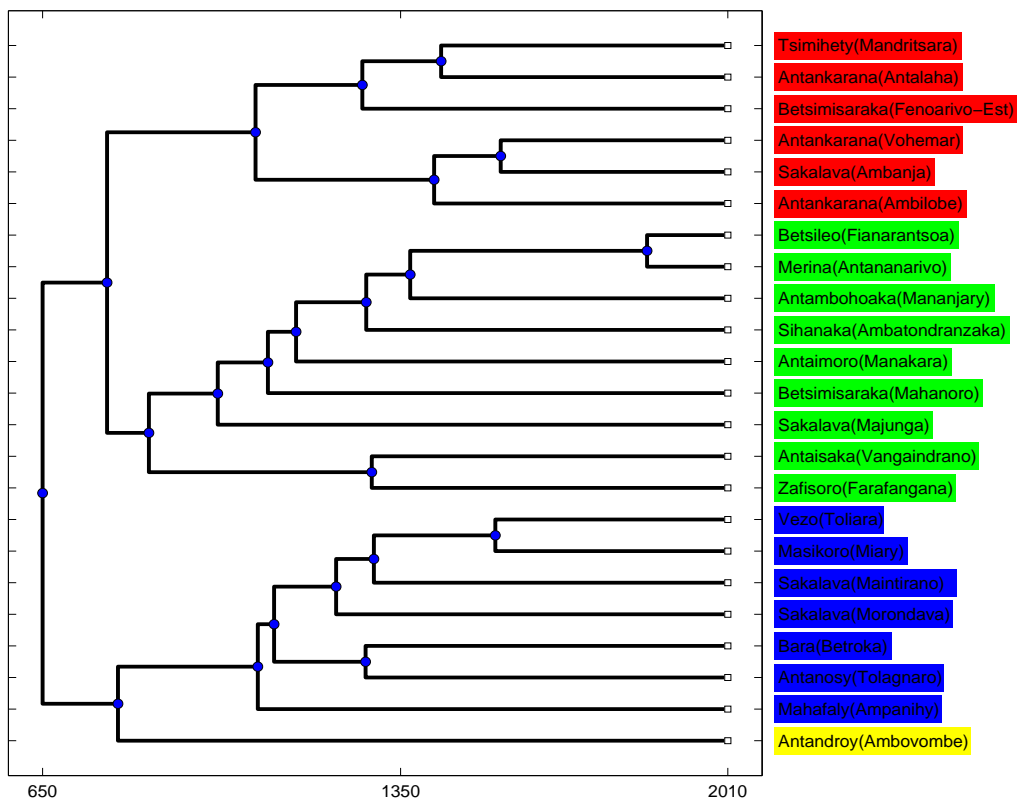, width=16cm, height =12cm}
\caption{\small UPGMA tree for 23 Malagasy dialects, with hypothetical 
separation times. Variants are named by traditional dialect names followed by 
locations in parentheses. The four main branches are colored distinctively.
The main separation of Malagasy dialects is
center-north-east vs. south-west. 
\label{Fig_1}}
\end{center}
\end{figure}

\begin{figure}[ht]
\begin{center}
\epsfig{file=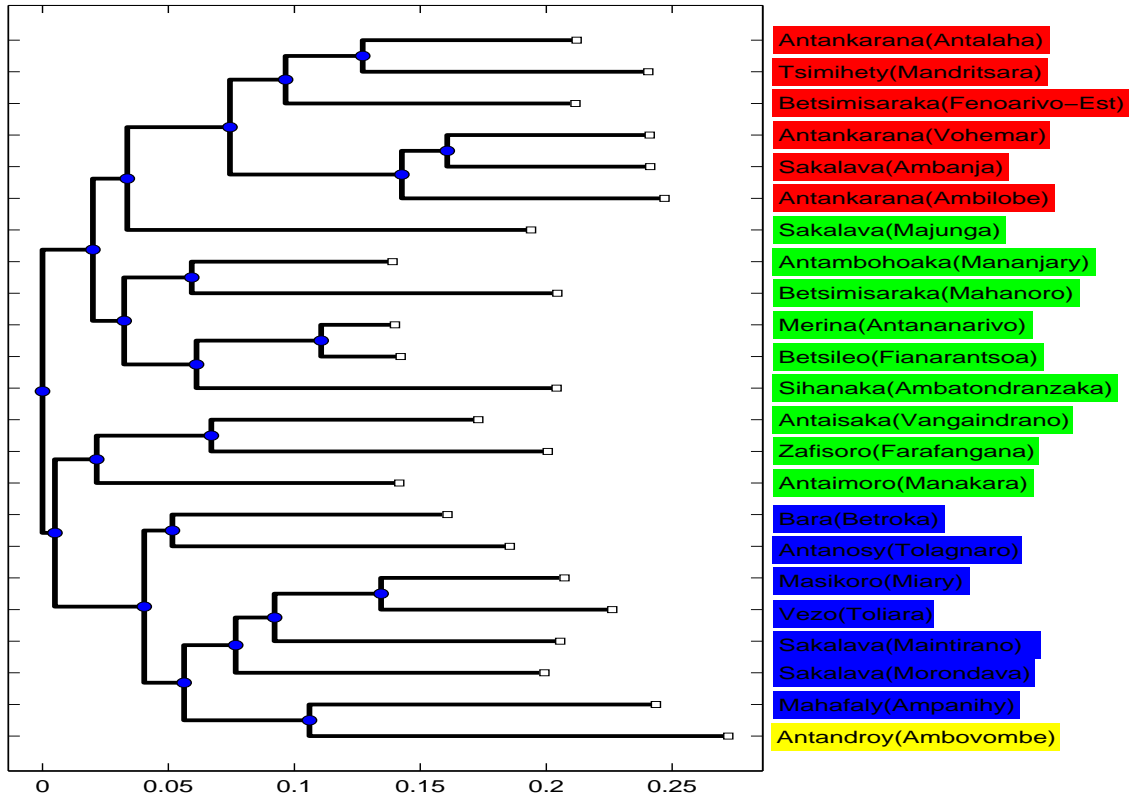, width=16cm, height =12cm}
\caption{\small NJ tree for 23 Malagasy dialects. Colors 
compare with the UPGMA tree in  Fig. \ref{Fig_1}.
The graph confirms the main center-north-east vs. south-west division.
The main difference is that three dialects at the 
linguistic border are grouped differently. Colors facilitate a rapid comparison.
\label{Fig_2}}
\end{center}
\end{figure}

\begin{figure}[ht]
\begin{center}
\epsfig{file=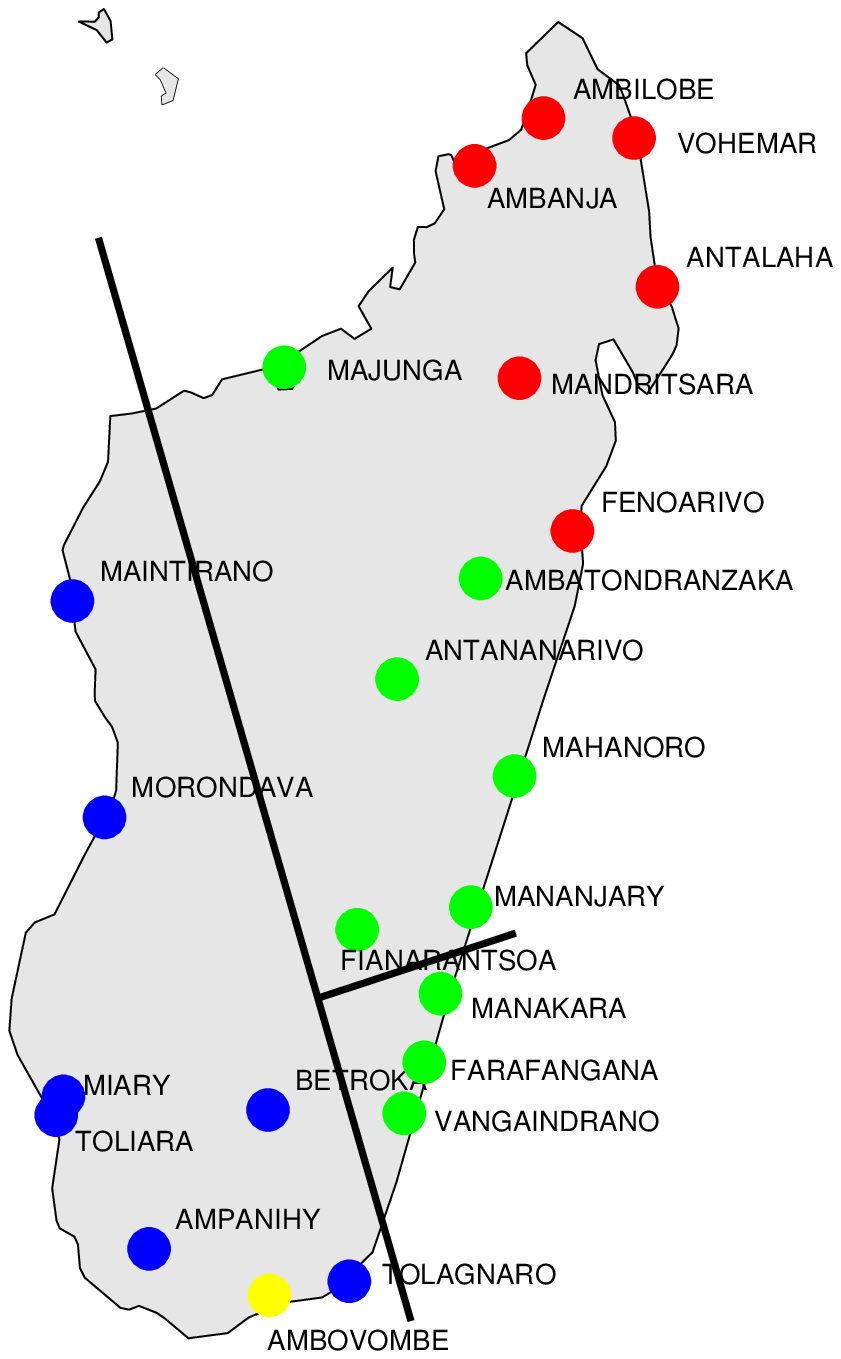, width=16cm, height =12cm}
\caption{\small Geographical locations of the 23 dialects studied, with colors
showing the main dialect branches according to Fig. \ref{Fig_1}. The straight line  
running from the south-east to the north-west of the island
corresponds to the basic split in UPGMA tree while the NJ grouping is similar
but Manakara, Farafangana and Vaingandrano are grouped with south-west. 
\label{Fig_3}}
\end{center}
\end{figure}

The two algorithms used are the Neighbor-Joining (NJ) \cite{Saitou:1987} and the
Unweighted Pair Group Method with Arithmetic Mean (UPGMA) \cite{Sokal:1958}. 
The main theoretical difference between the algorithms is that UPGMA assumes that 
evolutionary rates are the same on all branches of the tree, while NJ allows differences 
in evolutionary rates. The question of which method is better at inferring the phylogeny 
has been studied by running various simulations where the true phylogeny is known. 
Most of these studies were in biology but at least one \cite{Barbancon:2006} specifically 
tried to emulate linguistic data. Most of the studies (starting with \cite{Saitou:1987} 
and including \cite{Barbancon:2006}) found that NJ usually came closer to the 
true phylogeny. 
Since in our case, the relations among dialects are not necessarily
tree-like, it is desirable to test the different methods against empirical 
linguistic data, which is mainly why trees derived by means of both methods are presented here. 

The input data for the UPGMA tree are 
the pairwise separation times obtained from lexical distances by a rule
\cite{Serva:2008} which is a simple generalization of
the fundamental formula of glottochronology.
The absolute time-scale is calibrated by the results of the SCA analysis (see below), 
which indicate a separation date of A.D. 650.
While the scale below the UPGMA tree (Fig.~\ref{Fig_1}) refers to separation times, 
the scale below the NJ tree (Fig.~\ref{Fig_2}) simply shows lexical distance from the root.
The LDN distance between two language variants is roughly equal to the sum of their
lexical distance from their closest common node. 

Since UPGMA assumes equal evolutionary rates, the ends 
of all the branches line up on the right side of the UPGMA tree. The assumption of equal 
rates also determines the root of the tree on the left side. 
NJ allows unequal rates, so the ends of the branches do not all line up on the NJ tree. 
The extent to which they fail to line up indicates how variable the rates are. 
The tree is rooted by the midpoint (the point in the network equidistant from the two most 
distant dialects) but we also checked that the same result is obtained 
following the standard strategy of adding an out-group.

There is a good fit
between geographical position of dialects (see Fig. \ref{Fig_3}) and their position
both in the UPGMA (Fig. \ref{Fig_1}) and NJ trees (Fig. \ref{Fig_2}).
In both trees the dialects are divided into two main groups (colored blue and yellow vs.
red and green in Fig.~\ref{Fig_1}). 

Given the consensus between the two methods, the result regarding the 
basic split can be considered solid. Geographically the division corresponds to a border 
running from the south-east to the north-west of the island, as shown in Fig.~\ref{Fig_3}
where the UPGMA and NJ main separation lines are drawn.
A major difference concerns the Vangaindrano, Farafangana and Manakara dialects, 
which have shifting allegiances with respect to the two main groups under the different analyses. 
Additionally, there are minor differences in the way that the two main groups are configured 
internally. Most strikingly, we observe that in the UPGMA tree
Majunga is grouped with the central dialects while in the NJ tree it is grouped
with the northern ones. This indeterminacy would seem to relate to the fact that the town 
of Majunga is at the geographical border of the two regions.

Another difference is that in the UPGMA tree the Ambovombe variant 
of the dialect traditionally called Antandroy is quite isolated, whereas in the NJ tree 
Ambovombe and the Ampanihy variant of Mahafaly group together. 
Since the UPGMA algorithm is a strict bottom-up approach to the construction of a phylogeny, 
where the closest taxa are joined first, it will tend to treat the overall most deviant variant 
last. This explains the 
differential placement of Ambovombe in the two trees. 
The length of the branch leading to the node that joins Ambovombe and Ampanihy in the NJ 
tree shows that these two variants have quite a lot of similarities but 
in the UPGMA method these similarities in a sense 'drown' in 
the differences that set Ambovombe off from other Malagasy variants {\it as a whole}.

As a further confirmation of this analysis we also computed the average
LDN distance from each dialect to all the others.
Antandroy has the largest average distance, confirming that it is the overall most deviant 
variant (something which is also commonly pointed out by other Malagasy speakers). We further 
note that the smallest average distance is for the official variant, that of Merina. 
This may probably  be explained, at least in part, as an effect of the convergence of other 
variants towards this standard.

\subsection{\label{subsec:Their_results} The results of V\'{e}rin et al. (1969)}
\noindent
Our classification results, including the grouping of the dialects in a south-west and a 
center-east-north cluster, differ from \cite{Verin:1969}'s interpretation of their results, 
according to which there is a major split between the dialects on the northern tip of the 
island and all the rest. 

This divergence is somewhat surprising, so let us look into the way that V\'{e}rin 
{\it et al.} proceeded. 
There are some differences in the way that theirs and our datasets were 
constructed and the coverage. V\'{e}rin {\it et al.} used a 100-item Swadesh list, 
while we use a larger set of 200 words. We include locations that V\'erin {\it et al.}
did not cover. Moreover, following \cite{Gudschinsky:1956}, 
V\'{e}rin {\it et al.} (1969: 35) exclude Bantu loanwords from 
consideration, whereas we treat loanwords on a par with inherited words (in practice, 
however, V\'{e}rin {\it et al.}. only seem to identify one form as Bantu, 
namely {\it amboa} 'dog'.
Finally, a major difference is that V\'{e}rin {\it et al.} evaluated distances by 
the standard glottochronological approach based on cognate counting whereas we use the LDN 
measure.

\begin{figure}[ht]
\includegraphics[height=10cm,width=8cm]{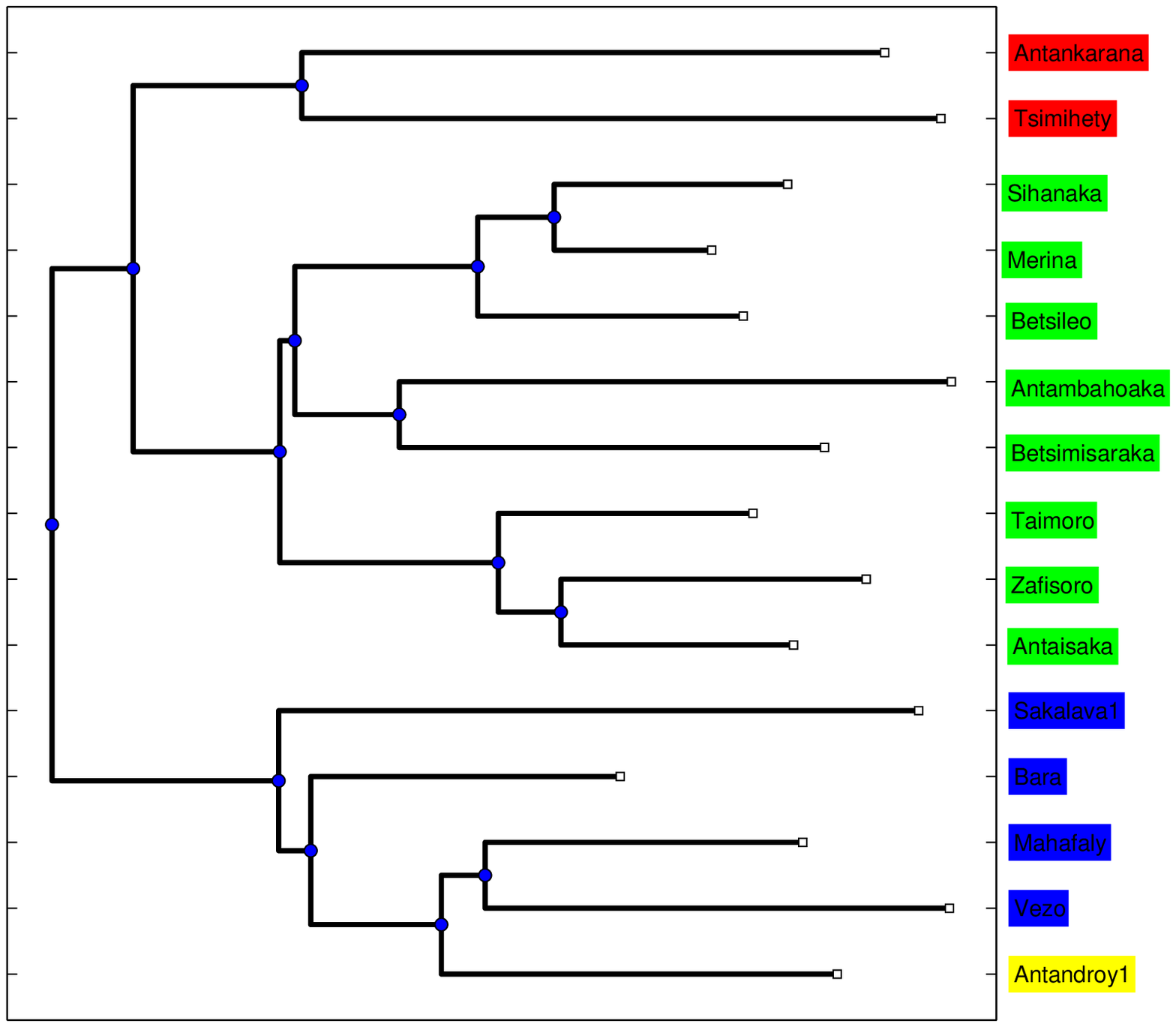}
\includegraphics[height=10cm,width=9cm]{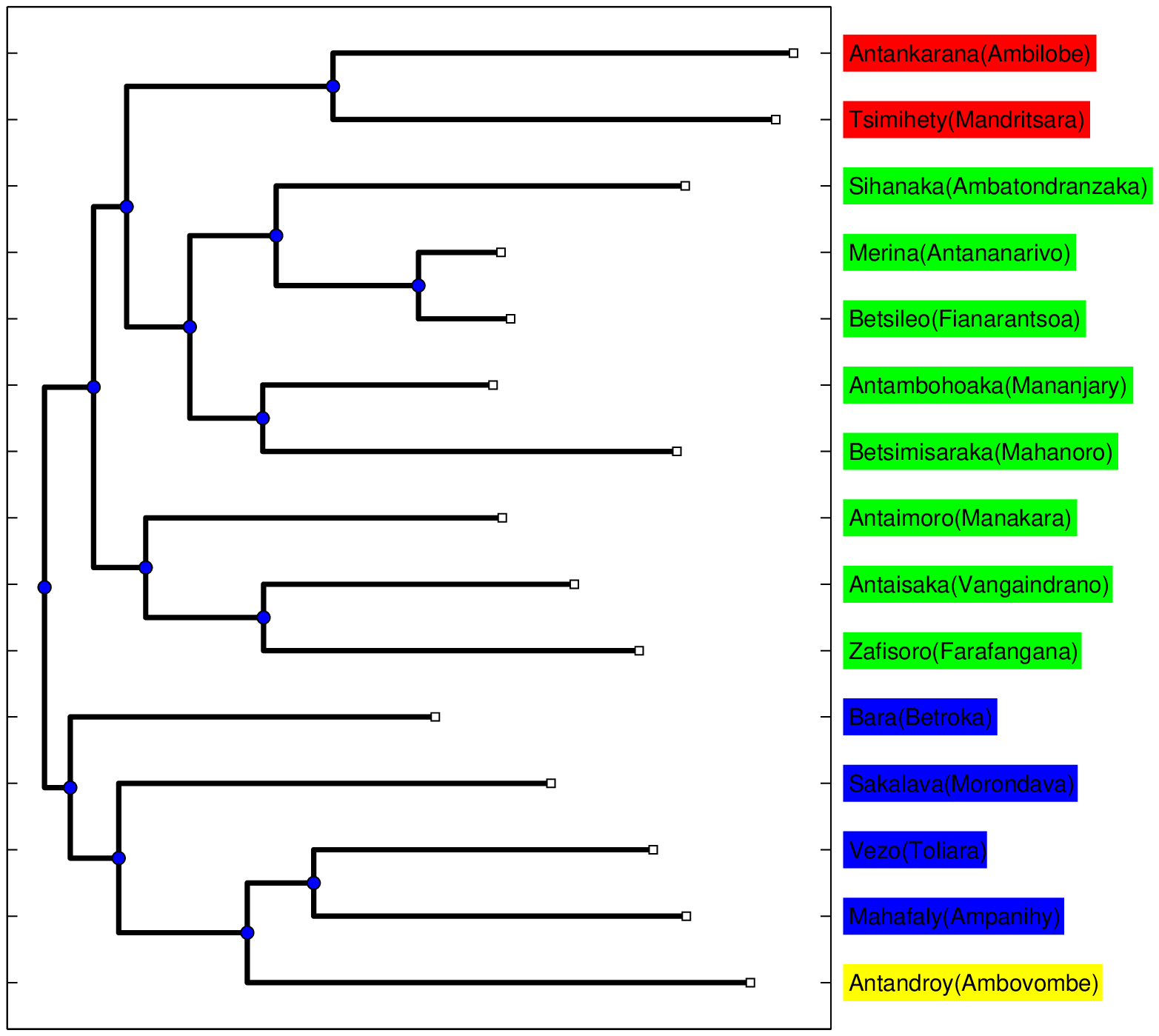}
\caption{\small Comparison between NJ trees based on respectively data collected by 
\cite{Verin:1969} (left) and by ourselves (right). 
The tree with V\'{e}rin data (Fig. 4a) is obtained by the standard lexicostatistical 
approach while the tree with our data (Fig. 4b) uses LDN distances.
Names of variants for the left tree are those of \cite{Verin:1969}, but the 
correspondence with our naming scheme, which makes use of to dialect names and towns, is evident. 
\label{Fig_4}}
\end{figure}

In spite of these differences our results are in reality quite similar to those 
of V\'{e}rin {\it et al.}, 
the differences mainly relating to the interpretation of their results. 
The great leap from results to interpretation is due to the fact that V\'{e}rin {\it et al.} 
did not have the kinds of sophisticated phylogenetic methods at their disposal for deriving 
a classification from a matrix of cognacy scores that are available today. 
Their method for constructing trees goes something like this: cluster the closest dialects 
first, using some threshold. Then move the threshold and join dialects or dialect groups 
under deeper nodes. Different trees can be constructed from using different thresholds. 
One of the problems with this approach, not addressed by the authors, 
is that it assumes a constant rate of change. 
For instance in one of their trees (their Chart 1 on p. 59) Merina, Sihanaka, and 
Betsileo Ambositra are joined under one node attached at the 92\% cognacy level. 
The actual percentages, however, do not fit a constant rate scenario 
(a.k.a. 'ultrametricality'): Sihanaka and Merina share 92\% cognates, 
Betsileo Ambositra and Merina also share 92\%, but Sihanaka and Betsileo Ambositra only 
share 86\%. No solution to this problem is given (and, indeed, it is a problem for any 
phylogenetic algorithm that cannot be 'solved' but at least needs to be addressed). 
Instead, violations of ultrametricality seem to be dealt with in an ad hoc way. 
In the case of the example just given Sihanaka and Betsileo Ambositra are treated as if 
they also shared 92\% cognates. Since the principles used by V\'erin {\it et al.}. to derive 
their trees are unclear there is no need to discuss their trees in detail. 
Moreover, each tree differs from the next, making it difficult to summarize the 
claims embodied in these trees. Some generalizations, however, do emerge. 
The Antankarana dialect in the far north constitutes its own isolated branch in all 
three trees, and in all three trees there are three sets of dialects that always belong 
together on different nodes: (1) Merina, Sihanaka, Betsileo Ambositra, Betsimisaraka; 
(2) Taimoro, Antaisaka, Zafisoro; (3) Mahafaly, Antandroy-1. 
Other dialects have no particularly close relationship to any other dialect, 
or else exhibit shifting allegiances. 

In Fig.~\ref{Fig_4}a we subject the distances data of V\'{e}rin {\it et al.} to NJ. 
Using this method each of the clusters (1-3) also turns up, but joined by other dialects 
which could not be safely placed at any deeper level of embedding by V\'{e}rin {\it et al.}. 
Thus, their clustering method essentially throws out so much information that only about 
half of the dialects become meaningfully classified. The most problematical aspect of 
their interpretation, however, is that there is supposed to be a fundamental split
between the Antankarana dialect in the far north of the island and all other dialects. 
As we demonstrate in Fig.~\ref{Fig_4}a, this is not borne out by the data, but is 
an artifact of the clustering method.

The NJ interpretation of the results of V\'{e}rin {\it et al.} (Fig.~\ref{Fig_4}a) may be 
compared to our own results obtained from the LDN distances evaluated using our own data 
(Fig.~\ref{Fig_4}b). 
Only variants belonging to the intersection of the two datasets are included. 
Names of variants are made identical, using the names from V\'{e}rin {\it et al.} 
The Betsimisaraka list from our data is the one from Mahanoro and the Antankarana list is 
the one from Vohemar. 

The two trees have similar topologies, in particular,
the main partition in both cases separates center-north-east from south-west 
dialects, which is at the variance the interpretation of V\'{e}rin {\it et al.} of 
their results. It is remarkable that the differences between the two trees are 
so minor considering differences both in the data 
and in the methods for 
calculating differences among dialects. 

Fig.~\ref{Fig_4}b was produced by using the same input
LDN distances and the same NJ algorithm as used for Fig.~\ref{Fig_2}. 
Comparing the two trees we observe that the simple reduction of
the number of input dialects has the effect of modifying the position of Farafangana,
Vankaindrano and Manakara variants (compare  Fig.~\ref{Fig_4}b with Fig.~\ref{Fig_2}). 
Indeed, the NJ tree in Fig.~\ref{Fig_4}b based on 15 dialects shows the same main branching 
as the UPGMA tree in Fig.~\ref{Fig_1}, which differs from that of the NJ tree in 
Fig.~\ref{Fig_2} based on 23 dialects.
This instability of tree topology caused by 
the number of input dialects and the differences in algorithms (UPGMA vs. NJ) shows that a 
tree structure
is not optimal for capturing all the information contained in the set of lexical distances.
Thus, we consider a different, geometrically-based approach, presented in the following section, 
necessary for a verification of classification results.

\section {\label{sec:Geometric_representation_of_Malagasy_dialects}
Geometric representation of Malagasy dialects}
\noindent
Although tree diagrams have become ubiquitous in representations of language taxonomies, 
they fail to reveal the full complexity of affinities among languages.
The reason is that the simple relation of ancestry, which is the single principle behind a 
branching family tree 
model, cannot grasp the complex social, cultural and political factors molding  
the evolution of languages \cite{Heggarty:2006}. 
Since all dialects within a group interact with each other 
and with the languages of other families in 'real time', it is obvious that any historical 
development in languages cannot be described only in terms of pair-wise interactions, 
but reflects a genuine higher order influence, which can best be assessed by Structural 
Component Analysis (SCA). This is a powerful tool which represents the relationships 
among different languages in a language family geometrically, in terms of distances 
and angles, as in the Euclidean geometry of everyday intuition. Being a version 
of the kernel PCA method \cite{Schoelkopf:1998}, it generalizes PCA to cases where we are 
interested in principal components obtained by taking all higher-order correlations 
between data instances. It has so far been tested through the construction of language 
taxonomies for fifty major languages of the Indo-European and Austronesian language 
families \cite{Blanchard:2010a}. The details of the SCA method are given in the Appendix B. 

In Fig.~\ref{Fig_5} we show the three-dimensional geometric representation of 23 
dialects of the Malagasy language and the Maanyan language, which is closely related to 
Malagasy.
The three-dimensional space is spanned by 
the three major data traits ($\{q_2,q_3,q_4\}$, see Appendix B for details) 
detected in the matrix of linguistic LDN distances.

\begin{figure}[ht]
\begin{center}
\epsfig{file=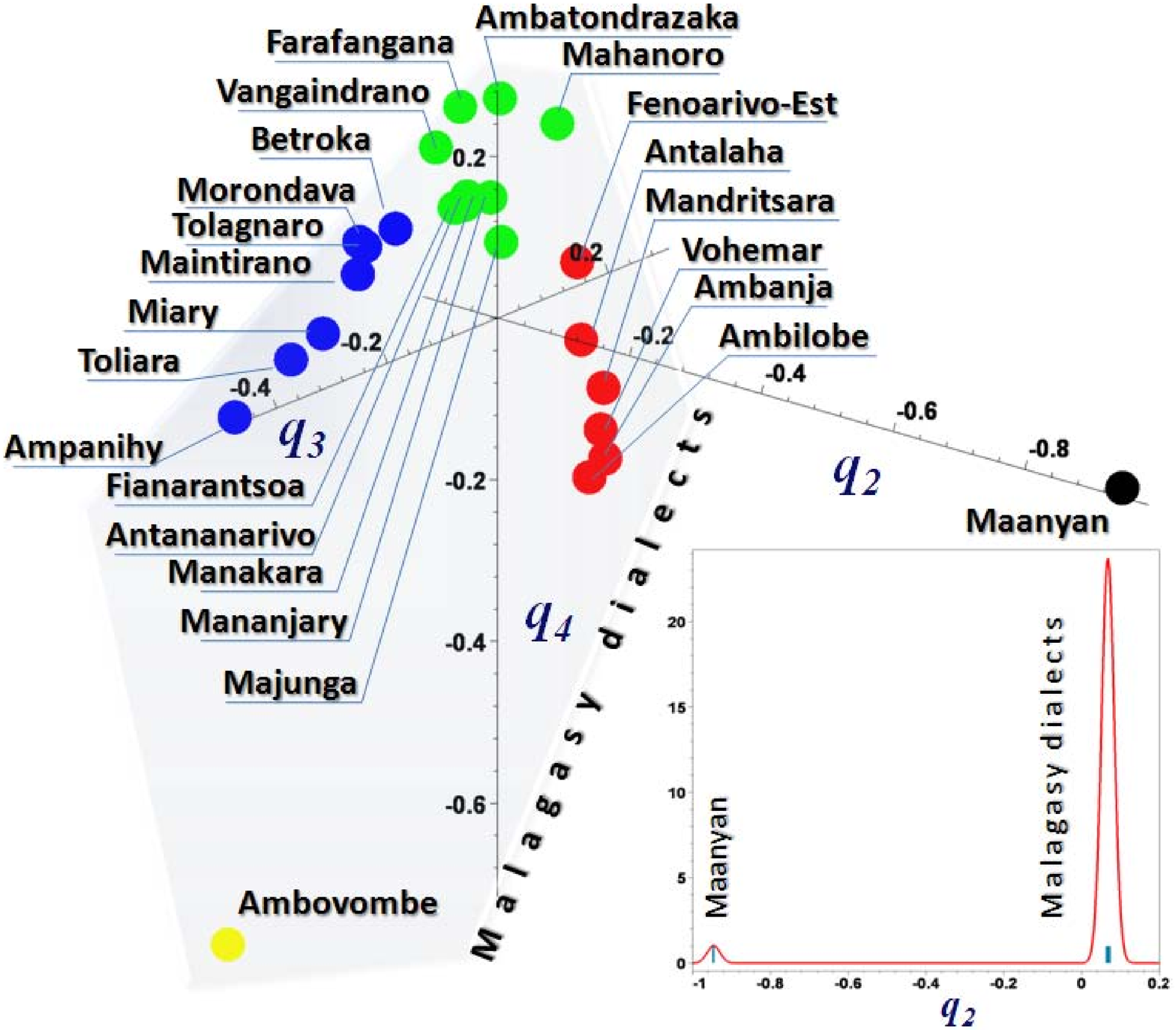, width=16cm, height =12cm}
\caption{\small The three-dimensional geometric representation of the Malagasy dialects
and the Maanyan language in the space of major data traits ($q_2, q_3, q_4$) 
shows a remarkable geographical patterning separating the northern (red) and the southern 
(blue) dialect groups, which fork from the central part of the island 
(the dialects spoken in the central part are colored green, while Antandroy is yellow).
The kernel density estimate of the distribution
of the $q_2$ coordinates, together with the absolute data frequencies,
indicate that all Malagasy dialects belong to a single plane 
orthogonal to the data trait of the Maanyan language ($q_2$).
\label{Fig_5}}
\end{center}
\end{figure}

\begin{figure}[ht]
\begin{center}
\epsfig{file=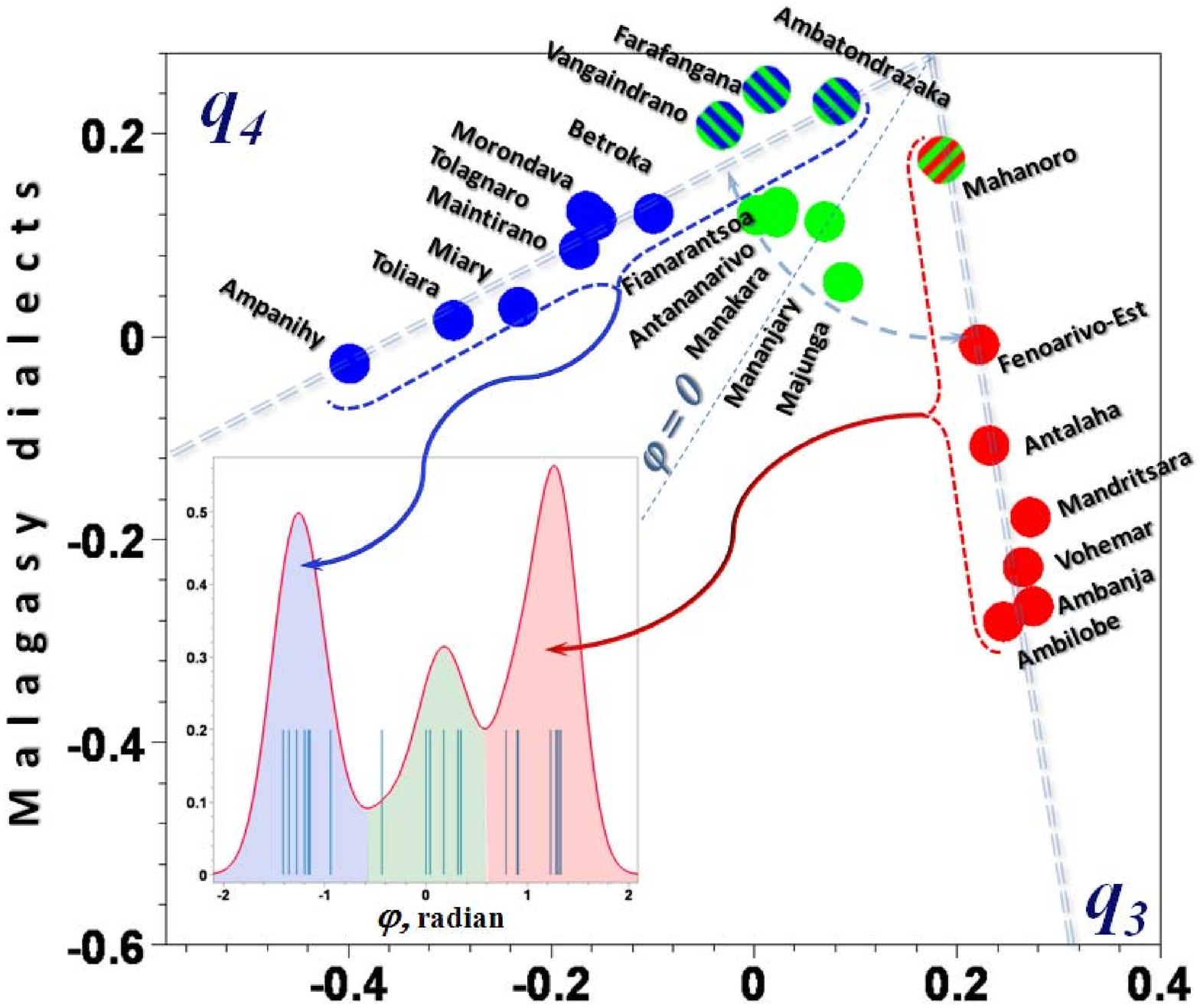,
 width=16cm, height =12cm}
\caption{\small The plane of Malagasy dialects ($q_3,q_4$);
Antandroy (Ambovombe) is excluded.
The kernel density estimate of the distribution over azimuth angles, together with the 
absolute data frequencies, allows the rest of Malagasy dialects to be classified into the three 
groups: north (red), south-west (blue), and center (green).
\label{Fig_6}}
\end{center}
\end{figure}

The clear geographical patterning is perhaps the most remarkable aspect of the geometric 
representation. The structural components reveal themselves in Fig.~\ref{Fig_5} as two 
well-separated spines representing both the northern (red) and the southern (blue) 
dialects of entire language. It is remarkable that all 
Malagasy dialects belong to a single plane orthogonal to the data trait of the Maanyan 
language ($q_2$). 
The plane of Malagasy dialects is attested by the sharp distribution of 
the language points in Cartesian coordinates along the data trait $q_2.$ 
This color point of Malagasy dialects over their common plane is shown in Fig.~\ref{Fig_6}
where a reference azimuth angle $\varphi$ is introduced in order to
underline the evident symmetry. 
It is important to mention that although the language point of Antandroy (Ambovombe) is 
located on the same plane as the rest of Malagasy dialects, it is situated far away from 
them and obviously belongs to neither of the dialect branches and for this reason is not 
reported in the next figure to be discussed (Fig.~\ref{Fig_6}).
This clear SCA isolation of Antrandroy is compatible with its position in the tree  
in  Fig.~\ref{Fig_1}.

The distribution of language points supports the main conclusion following from 
the UPGMA and NJ methods (Figs.~\ref{Fig_1}-\ref{Fig_2}) of a division of the 
main group of Malagasy dialects into three groups: north (red), south-west (blue) 
and center (green). These clusters are evident from the representation shown 
in Fig.~\ref{Fig_6}. However, with respect to the classification of some individual 
dialects the SCA method differs from the UPGMA and NJ results. 
Since their azimuthal coordinates better fit the general trend of the southern group, 
the Vangaindrano,  Farafangana, and  Ambatontrazaka dialects spoken in the central part 
of the island are now grouped with 
the southern dialects (blue) rather than the central ones 
(see Sec.~\ref{sec:The_internal_classification_of_Malagasy}). 
Similarly, the Mahanoro 
dialect is now classified in the northern group (red), since it is best fitted to the 
northern group azimuth angle. The remaining five dialects of the central group (green colored) 
are characterized by the azimuth angles close to a bisector ($\varphi=0$).

\begin{figure}[ht]
\begin{center}
\epsfig{file=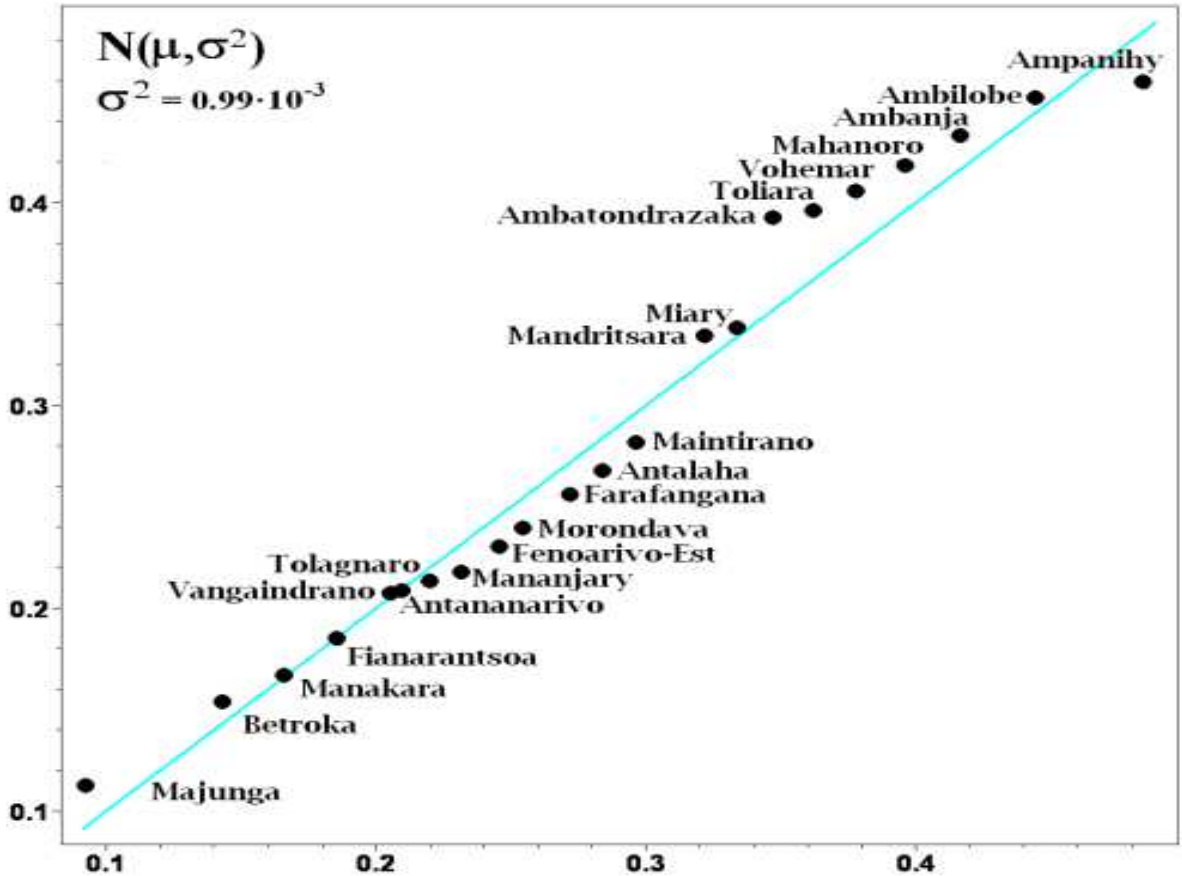,
 width=16cm, height =12cm}
\caption{\small The radial coordinates are ranked and then plotted
against their expected values under normality. Departures from linearity, 
which signify departures from normality, is minimal. 
\label{Fig_7}}
\end{center}
\end{figure}

\section{\label{sec:The_Arrival_to_Madagascar} The Arrival to Madagascar}

\subsection{\label{subsec:Dating} Dating the arrival}
\noindent
The radial coordinate of a dialect is simply the distance of
its representative point from the origin of coordinates in Fig.~\ref{Fig_5}.
It can be verified that the position of Malagasy dialects along the radial 
direction is remarkably heterogeneous indicating that the rates of change in the 
Swadesh vocabulary was anything but constant. 
  
The radial coordinates have been ranked and then plotted in Fig.~\ref{Fig_7} 
against their expected values under normality, such that departures from linearity 
signify departures from normality.
The dialect points in Fig.~\ref{Fig_7} show very good agreement with 
univariate normality with the value of variance $\sigma^2 = 0.99\times 10^{-3}$ 
which results from the best fit of the data. This normal behaviour
can be justified by the hypothesis that the dialect vocabularies are the result of a
gradual and cumulative process in which many small, independent innovations 
have emerged and to which they have additively contributed.

In the SCA method, which is based on the 
statistical evaluation of differences among the items of the Swadesh list, a complex nexus 
of processes behind the emergence and differentiation of 
dialects is described by the single degree of freedom (as another degree of freedom, 
the azimuth angle, is fixed by the dialect group) along the radial direction
\cite{Blanchard:2010b}.
 
The univariate normal distribution (Fig.~\ref{Fig_7}) 
implies a homogeneous diffusion time evolution in one dimension,  under which variance 
$\sigma^2\propto t$  grows linearly with time. The locations of dialect points would not 
be distributed normally if in the long run the value of variance $\sigma^2$ did not grow 
with time at an approximately constant rate. We stress that the constant rate of increase
in the variance of radial positions of languages  in the geometrical representation 
(Fig.~\ref{Fig_5}) has nothing to do with the traditional glottochronological assumption 
about the constant replacement rate of cognates assumed by the UPGMA method. 

It is also important to mention that the value of variance $\sigma^2= 0.99\times 10^{-3}$ 
calculated for 
the Malagasy dialects does not correspond to physical time but rather gives a statistically 
consistent estimate of age for the group of dialects. In order to assess the pace of 
variance changes with physical time and to calibrate the dating method we have used
historically attested events. Although the lack of documented historical events makes the 
direct calibration of the method difficult, we suggest (following \cite{Blanchard:2010a})
that variance evaluated over the Swadesh vocabulary proceeds approximately at the same 
pace uniformly for all human societies. For calibrating 
the dating mechanism in \cite{Blanchard:2010a}, we 
have used the following four anchoring historical events (see \cite{Fouracre:2007}) for the 
Indo-European language family: i.) the last Celtic migration (to the Balkans and Asia Minor) 
(by 300 BC); ii.) the division of the Roman Empire (by 500 AD); iii.) the migration of German 
tribes to the Danube River (by 100 AD); iv.) the establishment of the Avars Khaganate 
(by 590 AD) causing the spread of Slavic people. 
It is remarkable that all of the events mentioned uniformly indicate a very slow 
variance pace of a millionth per year, $t/\sigma^2=(1.367\pm 0.002)\cdot 10^6.$ 
This time-age ratio returns $t=1,353$ years if applied to the 
Malagasy dialects, suggesting that landing in Madagascar was around 650 A.D.
This is in complete agreement with the prevalent opinion among scholars 
including the influential one of Adelaar \cite{Adelaar:2009}.

\subsection{\label{subsec:The_landing_area} The landing area}
\noindent
In order to hypothetically infer the original center of dispersal of Malagasy variants,
we here use a variant of the method of \cite{Wichmann:2010a}. 
This method draws upon a well-known idea from biology \cite{Vavilov:1926} and linguistics 
\cite{Sapir:1916} that the homeland of a biological species or a language group corresponds 
to the current 
area of greatest diversity.
In \cite{Wichmann:2010a} this idea is transformed into quantifiable terms in the following way. 
For each language variant a diversity index is calculated as the average of the proportions 
between linguistic and geographical distances from the given language variant to each of 
the other language variants (cf. \cite{Wichmann:2010a} for more detail). 
The geographical distance is defined as the great-circle distance 
(i.e., as the crow flies) measured by angle radians. In this paper we adopt a variant of the 
method described in more detail in Appendix C.

The result of applying this method to Malagasy variants is that the best candidate for the 
homeland is the south-east coast where the three most diverse towns, i.e., Farafangana, 
Mahanoro and Ambovombe, are located, and where the surrounding towns are also highly diverse. 
The northern locations are the least diverse and they must have been settled last.

\begin{figure}[ht]
\begin{center}
\epsfig{file=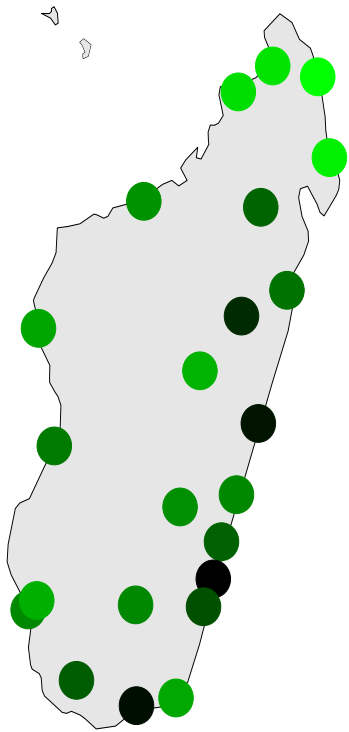,
 width=16cm, height =12cm}
\caption{\small  The homeland of Malagasy dialects as determined through diversity measures. 
The darkest-colored towns have the highest diversity values
while the light-colored the lowest. The most diverse area is the south-east coast where 
landing would have occurred, the less diverse area in the north, indicating that this 
area was settled last. 
\label{Fig_8}}
\end{center}
\end{figure}

A convenient way of displaying the results on a map is shown in Fig. \ref{Fig_8},
where location are indicated by means of 
circles with different color gradations. The greater
the diversity of a location is, the darker the color.
The figure suggest that the landing would have occurred somewhere between
Mahanoro (central part of the east cost) and Ambovobe (extreme 
south of the east coast), the most probable location being in the center of
this area, where Farafangana is situated.
Finally, we have checked that if the entire Greater Barito East group is considered, 
the homeland of Malagasy stays in the same place, but becomes secondary with respect to the 
south Borneo homeland of the group.

The identification of a linguistic homeland for Malagasy on the south-eastern coast of 
Madagascar receives some independent support from unexpected kinds of evidence. 
According to \cite{Faublee:1983} there is an Indian Ocean current that connects Sumatra with 
Madagascar. When Mount Krakatoa exploded in 1883, pumice was washed ashore on 
Madagascar's east coast where the Mananjary River opens into the sea
(between Farafangana and Mahanoro). During World War II 
the same area saw the arrival of pieces of wreckage from ships sailing between 
Java and Sumatra that had been bombed by the Japanese air-force. 
The mouth of the Mananjary River is where the town of Manajary is presently located, 
and it is in the highly diverse south-east coast
as shown in Fig. \ref{Fig_8}. To enter the current that would eventually 
carry them to the east coast of Madagascar the ancestors of today's Malagasy people would 
likely have passed by the easily navigable Sunda strait. 

In his studies on the roots of Malagasy, Adelaar finds that the 
language has an important contingent of loanwords from Sulawesi (Buginese)
\cite{Adelaar:1995b,Adelaar:2009}.
We have also compared Malagasy (and its dialects) with various Indonesian languages. 
While we unsurprisingly find that Maanyan is the closest language, we also find that
Buginese is the third closest one (see also \cite{Petroni:2008}).
The similarity with Buginese appears to be a further argument in favor of the southern 
path through the Sunda strait to Madagascar. In fact, if the Malay sailors 
recruited their crew in Borneo and, at a limited extent, in Sulawesi,
they likely crossed this strait before starting their navigation in the open waters.

Furthermore, we found that the dialects of Manajary, Manakara, Antananarivo and Fianarantsoa are
noticeably closer both to Maanyan and Malay with respect to the other variants.
Manajary and Manakara are both in the identified landing area in the south-east coast while 
Antananarivo and Fianarantsoa are in the central highlands of Madagascar. This fact may
suggest that landing was followed shortly after by a migration to the interior of the Island.

\section{\label{sec:Conclusion} Conclusion and outlook}
\noindent

All results in this paper rely on two main ingredients: a new dataset from 23 different 
variants of the language and an automated method to evaluate lexical distances.
Analyzing the distances through different types of phylogenetic algorithm (NJ and UPGMA) 
as well as through a geometrical approach we find that all approaches converge on a result 
where dialects are classified into two main geographical 
subgroups: south-west vs. center-north-east.
It is not clear, at this stage, if this main division is caused by geography or by an early 
splitting of the population into two different subpopulations or even by a colonization history 
with more than one founding nucleus. The last hypothesis, however, 
is somewhat unlikely given the relative uniformity of the dialects.

An output of the geometric representation of the distribution of the dialects is a landing
date of around 650 AD, in agreement with a view commonly held by students of Malagasy.
Furthermore, by means of a technique which is based on the calculation of differences in 
linguistic 
diversity, we propose that the south-east coast was the location were the first colonizers 
landed. This location also suggests that the path followed by the sailors 
went from Borneo, through the Sunda strait, and subsequently, along major oceanic currents, 
to Madagascar.

Finally, we measured the distance of the Malagasy variants to other Indonesian languages 
and found that the dialects of Manajary, Manakara, Antananarivo and Fianarantsoa are 
noticeably closer to most of them.

A larger comparison of Malagasy variants with Indonesian (and possibly African) 
languages is desirable. 
Although Malagasy is assigned to the the Greater East Barito group 
it has many loanwords from other Indonesian languages such as Javanese, Buginese, and Malay, 
especially in the domains of maritime life and navigation.
It has also been observed that it is unlikely that Maanyan speaking Dayaks were responsible 
for the spectacular migrations from Kalimantan to Madagascar since they are forest dwellers 
with river navigation skills only.
Furthermore, many manifestations of Malagasy culture cannot be linked up with the culture of 
the Dayaks of the south-east Barito area. 
For example, the Malagasy people use outrigger canoes, whereas south-east Barito Dayaks 
never do; some of the Malagasy musical instruments are very similar to musical instruments 
in Sulawesi; and some of the Malagasy cultigens (wet rice) cannot be found among Barito 
river inhabitants. 
In contrast, some funeral rites, such as the {\it famadihana} (second burial), 
are similar to those of Dayaks.
Nevertheless, it should be observed that it is not clear whether the above cultural traits 
are specific to a region or a people or whether they are generic traits that can be 
found sporadically in other Austronesian cultures.

Non-Maanyan cultural and linguistic traits
raise several questions concerning the ancestry of the Malagasy people.
Assuming that Dayaks were brought as subordinates together with a few other Indonesians by 
Malay seafarers, they formed the majority in the initial group and their language constituted 
the core element of what later became Malagasy. 
In this way Malagasy would have absorbed words of the Austronesian languages of the other slaves 
and of the Malay seafarers.
Is this a  sufficient explanation, or are things are more complicated? 
For example, may we hypothesize two or more founding colonies with different ethnic 
composition? And is it possible that
later specific contacts altered the characteristics of some local dialects?

In order to answer these questions it is necessary to make a careful comparison of 
all Malagasy variants with all Austronesian languages. 
A dialect may provide information about the pre-migratory composition of its speakers and also 
about further external contributions due to successive landing of Indonesian sailors.

Furthermore, the island was almost surely inhabited before the arrival of Malagasy ancestors.
Malagasy mythology portrays a people, called the {\it Vazimba}, as the original inhabitants, 
and it is not clear whether they were part of a previous Austronesian expansion or a population
of a completely different origin (Bantu, Khoisan?).
Is it possible to track the aboriginal vocabulary into the some of the dialects such as, 
for example, Mikea (see \cite{Blench:2010})? 

These questions call for a new look at the Malagasy language, not as a single entity, 
but as a constellation of variants  whose histories are still to be fully understood.

\section*{\label{sec:Acknowledgements} Acknowledgments}
\noindent
We thank Philippe Blanchard and Eric W. Holman for comments on an earlier version of this paper. 
Further, M. Serva wishes to thank Sanhindou Amady, Clement 'Zazalahy', Beatrice Rolla, 
Renato Magrin, Corto Maltese and Gianni Dematteo for logistical support during his 
stay in Madagascar.

\newpage

\section*{\label{sec:Appendix_A} Appendix A}
\noindent 
The lexical distance \cite{Serva:2008, Petroni:2008, Holman:2008} between the two 
languages,  $l_i$ and $l_j$, is computed as the average of the normalized Levenshtein 
(edit) distance 
\cite{Levenshtein:1966} 
over the vocabulary of 200 items,
\begin{equation}
D\left(l_i,l_j\right)\,\,=\,\, \frac{1}{200}\sum_{\alpha=1}^{200}
\frac{\left\| w_i (\alpha),w_j (\alpha) \right\|}
{\max \left( \left| w_i (\alpha) \right|,\left| w_j (\alpha)\right|
\label{distance} 
\right)}.
\end{equation}

\noindent
where the item is indicated by $\alpha$,
 $\left\| w_i(\alpha),w_j(\alpha) \right\|$ is the standard Levenshtein distance 
between the words $w_i(\alpha)$ and $w_j(\alpha)$, 
and $|w_i(\alpha)|$ is the number of characters in the word $w_i(\alpha)$ . 
The sum runs over all the 200 different items of the Swadesh list.
Assuming that the number of languages (or dialects) 
to be compared is $N$, then the
distances $D(l_i,l_j)$  are the entries of a $N \times N$ symmetric matrix ${\bf D}$
(obviously $D(l_i,l_i)=0$).

The matrix, with entries multiplied by 1000, is the following:
\bigskip
\bigskip
 
\small
\setlength{\tabcolsep}{3.2pt}
\begin{tabular}{l | l*{22}{c}r}
 1 \\
 2 & 323 \\
 3 & 246 & 276 \\
 4 & 322 & 240 & 295 \\
 5 & 302 & 281 & 309 & 345 \\
 6 & 227 & 318 & 275 & 359 & 266 \\
 7 & 413 & 386 & 390 & 418 & 314 & 370 \\
 8 & 280 & 386 & 342 & 401 & 356 & 245 & 436 \\
 9 & 366 & 424 & 379 & 412 & 405 & 375 & 450 & 409 \\
10 & 411 & 396 & 416 & 440 & 318 & 366 & 249 & 456 & 482 \\
11 & 207 & 326 & 260 & 362 & 286 & 061 & 383 & 201 & 374 & 384 \\
12 & 362 & 343 & 345 & 387 & 292 & 328 & 289 & 397 & 435 & 330 & 324 \\
13 & 303 & 369 & 330 & 381 & 384 & 329 & 454 & 362 & 256 & 487 & 318 & 407 \\
14 & 343 & 302 & 331 & 355 & 243 & 317 & 303 & 403 & 423 & 314 & 336 & 301 & 419 \\
15 & 397 & 453 & 394 & 462 & 392 & 375 & 342 & 463 & 485 & 304 & 383 & 405 & 471 & 388 \\
16 & 368 & 391 & 385 & 416 & 392 & 390 & 448 & 406 & 320 & 474 & 383 & 429 & 325 & 418 & 486 \\
17 & 400 & 350 & 369 & 390 & 280 & 358 & 165 & 433 & 427 & 278 & 373 & 240 & 439 & 261 & 358 & 410 \\
18 & 322 & 376 & 325 & 374 & 391 & 337 & 426 & 381 & 198 & 473 & 339 & 412 & 234 & 406 & 461 & 264 & 414 \\
19 & 358 & 407 & 376 & 417 & 408 & 394 & 440 & 419 & 292 & 481 & 387 & 431 & 325 & 422 & 472 & 161 & 408 & 243 \\
20 & 297 & 388 & 359 & 430 & 356 & 299 & 400 & 346 & 386 & 433 & 275 & 375 & 363 & 375 & 455 & 348 & 394 & 349 & 355 \\
21 & 386 & 341 & 370 & 385 & 290 & 344 & 262 & 403 & 422 & 321 & 348 & 250 & 404 & 306 & 403 & 401 & 213 & 416 & 417 & 383 \\
22 & 225 & 389 & 332 & 394 & 382 & 316 & 471 & 319 & 385 & 475 & 287 & 421 & 296 & 431 & 480 & 382 & 467 & 348 & 387 & 356 & 441 \\
23 & 379 & 424 & 407 & 424 & 398 & 380 & 443 & 433 & 315 & 466 & 380 & 412 & 351 & 420 & 472 & 203 & 395 & 288 & 202 & 351 & 409 & 406 \\
\hline
 & 1 & 2 & 3 & 4 & 5 & 6 & 7 & 8 & 9 & 10 & 11 & 12 & 13 & 14 & 15 & 16 & 17 & 18 & 19 & 20 & 21 & 22 \\

\end{tabular}

\normalsize
\bigskip
\bigskip
\noindent
where the number-variant correspondence is:

\noindent
1 Antambohoaka (Mananjary), 
2 Antaisaka (Vangaindrano), 
3 Antaimoro (Manakara), 
4 Zafisoro (Farafangana), 
5 Bara (Betroka), 
6 Betsileo (Fianarantsoa), 
7 Vezo (Toliara), 
8 Sihanaka (Ambatondranzaka), 
9 Tsimihety (Mandritsara), 
10 Mahafaly (Ampanihy), 
11 Merina (Antananarivo), 
12 Sakalava (Morondava), 
13 Betsimisaraka (Fenoarivo-Est), 
14 Antanosy (Tolagnaro), 
15 Antandroy (Ambovombe), 
16 Antankarana (Vohemar), 
17 Masikoro (Miary), 
18 Antankarana (Antalaha), 
19 Sakalava (Ambanja), 
20 Sakalava (Majunga), 
21 Sakalava (Maintirano), 
22 Betsimisaraka (Mahanoro), 
23 Antankarana (Ambilobe).

\section*{\label{sec:Appendix_B} Appendix B}
\noindent 
The lexical distance (\ref{distance}) between two languages, $l_i$ 
and $l_j$, can be interpreted as the average probability to distinguish them by a mismatch 
between two characters randomly chosen from the orthographic realizations of the vocabulary 
meanings. There are infinitely many matrices that match all the structure of ${\bf D}$, 
and therefore contain all the information about the relationships between languages, 
\cite{Blanchard:2010a}. It is remarkable that all these matrices are related to each 
other by means of a linear transformation,
\begin{equation}
\label{random_walk}
{\bf T }\,\,=\,\,
{\bf \Delta}^{-1} {\bf D },\quad 
{\bf \Delta}=\mathrm{diag}\left(
\sum_{k=1}^N D\left(l_1,l_k\right)\ldots
 \sum_{k=1}^N D\left(l_N,l_k\right)
\right),
\end{equation}
which can be interpreted as a random walk \cite{Blanchard:2010a,Blanchard:2010b} 
defined on the weighted undirected graph determined by the matrix of lexical distances 
${\bf D}$ over the $N$ different languages. Random walks defined by the transition matrix 
(\ref{random_walk}) describe the statistics of a sequential process of language 
classification. Namely, while the elements $T(l_i, l_j)$ of the matrix ${\bf T}$ evaluate 
the probability of successful differentiation of the language $l_i$ provided the 
language $l_j$ has been identified certainly, the elements of 
the squared matrix ${\bf T}^2$, 
ascertain the successful differentiation of the language $l_i$ from $l_j$ through an 
intermediate language, the elements of the matrix ${\bf T}^3$ give the probabilities to 
differentiate the language through two intermediate steps, and so on. 
The whole host of complex and indirect relationships between orthographic 
representations of the vocabulary meanings encoded in the matrix of lexical 
distances (\ref{distance}) is uncovered by the von Neuman series estimating the 
characteristic time of successful classification for any two languages in the database 
over a language family,\begin{equation}
\label{kernel}
{\bf J}\,\,=\,\,
\lim_{n\to\infty}
\sum_{k=0}^n {\bf T}^n\,\,=\,\, \frac 1{1-{\bf T}}.
\end{equation}
The last equality in (\ref{kernel}) is understood as the group generalized inverse 
(Blanchard:2010b) being a symmetric, positive semi-definite matrix which plays the 
essentially same role for the SCA, as the covariance matrix does for the usual PCA analysis. 
The standard goal of a component analysis (minimization of the data redundancy quantified by 
the off-diagonal elements of the kernel matrix) is readily achieved by solving an eigenvalue 
problem for the matrix ${\bf J}$. Each column vector $q_k$, which determines a direction 
where $\bf J$ acts as a simple rescaling, ${\bf J}q_k = \lambda_kq_k$, with some real 
eigenvalue $\lambda_k = 0$, is associated to the virtually independent trait in the matrix 
of lexical distances ${\bf D}$. Independent components 
$\left\{q_k\right\}$, $k = 1,\ldots N$, define an orthonormal basis in $\mathbb{R}^N$ 
which specifies each language $l_i$ by $N$ numerical coordinates, 
$l_i \to \left(q_{1,i}, q_{2,i},\ldots q_{N,i}\right)$. 
Languages that cast in the same mold in accordance with the $N$ individual 
data features are revealed by geometric proximity in Euclidean space spanned by the 
eigenvectors $\left\{q_k\right\}$ that might be either exploited visually, or accounted 
analytically. The rank-ordering of data traits $\left\{q_k\right\}$, in accordance to their 
eigenvalues, $\lambda_0 =\lambda_1 <\lambda_2 =\ldots= \lambda_N$, provides us with the
 natural geometric framework for dimensionality reduction. At variance with the standard 
PCA analysis \cite{Jolliffe:2002}, where the largest eigenvalues of the covariance matrix 
are used in order to identify the principal components, while building language taxonomy, 
we are interested in detecting the groups of the most similar languages, with respect to 
the selected group of features. The components of maximal similarity are identified with the 
eigenvectors belonging to the smallest non-trivial eigenvalues. Since the minimal 
eigenvalue $\lambda_1 = 0$ corresponds to the vector of stationary distribution of random 
walks and thus contains no information about components, we have used the three consecutive 
components $\left(q_{2,i}, q_{3,i},q_{4,i}\right)$ as the three Cartesian coordinates of a 
language $l_i$ in order to build a three-dimensional geometric representation 
of language taxonomy. Points symbolizing different languages in space of the three major 
data traits are contiguous if the orthographic representations of the vocabulary meanings 
in these languages are similar.

\section*{\label{sec:Appendix_C} Appendix C}
\noindent 
The lexical distance $D(l_i,l_j)$ between two dialects $l_i$ and $l_j$
was previously defined; their geographical distance $\Delta(l_i,l_j)$  can be simply
defined as the distance between the two locations where the dialects were collected.
There are different possible measure units for $\Delta(l_i,l_j)$.
We simply use the great-circle angle (the angle that the two location form with the center 
of the earth).

It is reasonable to assume, in general, that larger geographical distances
correspond to larger lexical distances and vice-versa
For this reason in \cite{Wichmann:2010a} the diversity was measured 
as the average of the ratios between lexical and geographical distance.
This definition implicitly assumes that lexical distances vanish
when geographical distances equal 0. Nevertheless, different dialects 
are often spoken at the same locations, separated by negligible geographical
distances. For this reason, and because a zero denominator in the division involving
geographical distances would cause some diversity indexes to become infinite,
\cite{Wichmann:2010a} arbitrarily added a constant of .01 km to all distances. 

Here we similarly add a constant, but one whose value is better motivated.
We plotted all the $\frac {23 \times 22}{2}=253$ points 
$\Delta(l_i,l_j), \, D(l_i,l_j)$ in a bi-dimensional space
and verified that the pattern is compatible with a linear shape
in the domain of small geographical distances.
Linear regression of the $20\%$ of points with smaller 
geographical distances  gives the interpolating straight line $D=a+b\Delta$ with 
$a=0.22$ and $b=0.04$.
The results indicates that a lexical distance of $0.22$ is 
expected between two variants of a language spoken in coinciding locations. 

The choice of constants $a$ and $b$ by linear regression
assures that the ratio between $D(l_i,l_j)$ and $a + b\Delta(l_i,l_j)$ is around 1
for any pair of dialects  $l_i$ and $l_j$. 
A large value of the ratio corresponds to a pair of variants which are lexically 
more distant and vice-versa.
It is straightforward to define the diversity of a dialect as 

\begin{equation}
V(l_i) \,\, = \,\, \frac{1}{22} \sum_{j \neq i}
\frac{D(l_i,l_j)}{a + b\Delta(l_i,l_j)}
\label{diversity}.
\end{equation}
in this way, locations with high diversity will be characterized by a a larger $V(l_i)$,
while locations with low diversity will have a smaller one.

Notice that the above definition coincides with the one in \cite{Wichmann:2010a},
the main difference being that instead of an arbitrary value of $a$ we obtain
it through the output of linear regression. 

The diversities (in a decreasing order), computed with (\ref{diversity}), are the following:
Zafisoro (Farafangana): 1.00,  
Betsimisaraka (Mahanoro): 0.98,
Antandroy (Ambovombe): 0.98, 
Sihanaka (Ambatontrazaka): 0.95,
Antaisaka (Vangaindrano): 0.92,  
Mahafaly (Ampanihy): 0.90, 
Tsimihety (Mandritsara): 0.90, 
Antaimoro (Manakara): 0.90, 
Betsimisaraka (Fenoarivo-Est): 0.88, 
Sakalava (Morondava): 0.87, 
Antambohoaka (Mananjary): 0.86,
Vezo (Toliara): 0.86, 
Bara (Betroka): 0.86,
Sakalava (Majunga): 0.85,
Betsileo (Fianarantsoa): 0.85,
Antanosy (Tolagnaro): 0.83, 
Sakalava (Maintirano): 0.83, 
Masikoro (Miary): 0.82,
Merina (Antananarivo): 0.82, 
Sakalava (Ambanja): 0.77, 
Antankarana (Ambilobe): 0.77,
Antankarana (Antalaha): 0.76,
Antankarana (Vohemar): 0.74.

\section*{\label{sec:Appendix_D} Appendix D}
\noindent
In Table 1 below we provide information on the people who furnished the data collected by 
one of us (M.S.) at the beginning of 2010 with the invaluable help of 
 Joselin\`a Soafara N\'{e}r\'{e}. For any dialect (except for Merina,
for which published lists combined with the personal knowledge of M.S. were used), 
data were elicited independently from two consultants.
Their names and birth dates follow each of the dialect names.

\begin{table}[p]
\caption{People who
 furnished the data on Malagasy dialects \label{Tab:1}}
\centering
\begin{tabular}[ht]{l l l}
\hline
\small \bf MERINA &\small  & \small \\
\small \bf(ANTANANARIVO)       
             & \small SERVA Maurizio &\small   \\ \hline
\small\bf ANTANOSY
             &  \small SOAFARA Joselina Nere & 
                                         \small  08 November 1987 \\
\small\bf  (TOLAGNARO)   &\small ETONO Imasinoro Lucia &\small  18 February 1982\\  \hline
\small\bf BETSIMISARAKA &
\small ANDREA Chanchette G\'{e}n\'{e}viane &\small  07 August 1985 \\
\small \bf  (FENOARIVO-EST)  &\small RAZAKAMAHEFA Joachim Julien   & \small 09 November 1977\\
\hline
\small \bf SAKALAVA &
 \small SEBASTIEN Doret & \small 26 November 1980 \\
 \small \bf  (MORONDAVA)  &\small RATSIMANAVAKY Christelle J.    & \small 29 February 1984\\
\hline
\small \bf VEZO &
\small RAKOTONDRABE Justin &\small  02 August 1972 \\
 \small \bf  (TOLIARA)  &RASOAVAVATIANA Claudia S.    &  \small 28 June 1983\\
\hline
\small\bf ZAFISIRO &
\small RALAMBO Alison & \small 11 June 1982 \\
\small \bf  (FARAFANGANA)  & \small RAZANAMALALA Jeanine    &\small  03 February 1980\\
\hline
\small \bf ANTAIMORO &
\small RAZAFENDRALAMBO Haingotiana & \small 24 July 1985 \\
\small \bf  (MANAKARA)  &\small RANDRIAMITSANGANA Blaise    & \small  05 February 1989\\
\hline
\small \bf ANTAISAKA &
\small RAMAHATOKITSARA Fidel Justin &\small  24 April 1984\\
\small \bf  (VANGAINDRANO)  & \small FARATIANA Marie Luise    &\small   17 August 1990\\
\hline
\small \bf ANTAMBOHOAKA &
\small RAKOTOMANANA Roger & \small 04 May 1979\\
\small  \bf  (MANANJARY)  &\small ZAFISOA Raly    & \small   20 April 1983\\
\hline
\small\bf BETSILEO &
\small RAMAMONJISOA Andrininina Leon Fidelis &\small 16 April 1987\\
 \small \bf  (FIANARANTSOA)  & \small RAKOTOZAFY Teza   &  \small 25 December 1985\\
\hline
\small\bf BARA &\small
RANDRIANTENAINA Hery Oskar Jean &\small 17 Jenuary 1986\\
\small \bf  (BETROKA)  & \small NATHANOEL Fife Luther  & \small  26 May 1983\\
\hline
\small\bf TSIMIHETY &
\small RAEZAKA Francis & \small 23 December 1984\\
\small \bf  (MANDRITSARA)  &\small FRANCINE Germaine Sylvia  &  \small 04 May 1985\\
\hline
\small \bf MAHAFALY &
\small VELONJARA Larissa &\small 21 April 1989\\
 \small \bf  (AMPANIHY)  &\small NOMENDRAZAKA Christian  & \small  07 June 1982\\
\hline
\small \bf SIHANAKA &
\small ARINAIVO Robert Andry & \small 06 Jenuary 1979\\
 \small \bf  (AMBATONDRAZAKA)  & \small RONDRONIAINA Natacha  &  \small 27 December 1985\\
\hline
\small\bf ANTANKARANA &
\small ANDRIANANTENAINA N. Benoit& \small 06 August 1984\\
\small \bf  (VOHEMAR)  &\small EDVINA Paulette & \small  28 Jenuary 1982\\
\hline
\small\bf ANTANKARANA &
\small RANDRIANARIVELO Jean Ives& \small 24 December 1986\\
\small \bf  (ANTALAHA)  &\small RAZANAMIHARY Saia &
 \small   07 September 1985\\
\hline
\small\bf SAKALAVA &
\small CASIMIR Jaozara Pacific& \small 03 April 1983\\
\small \bf  (AMBANJA)  &\small ZAKAVOLA M. Sandra &
 \small   17 July 1984\\
\hline
\small\bf SAKALAVA &
\small RATSIMBAZAFY Serge& \small 17 May 1978\\
\small \bf  (MAJUNGA)  &\small VAVINIRINA Fideline &
 \small    23 June 1970\\
\hline
\small\bf ANTANDROY &
\small RASAMIMANANA Z. Epaminodas& \small 05 June 1983\\
\small \bf  (AMBOVOMBE)  &\small MALALATAHINA Tiaray Samiarivola &
 \small    07 July 1984\\
\hline
\small\bf MASIKORO &
\small MAHATSANGA Fitahia& \small 22 March 1976\\
\small \bf  (ANTALAHA)  &\small VOANGHY Sidonie Antoinnette &
 \small    12 October 1981\\
\hline
\small\bf ANTANKARANA &
\small BAOHITA Maianne& \small 21 August 1984\\
\small \bf  (AMBILOBE)  &\small NOMENJANA HARY Jean Pierre Felix &
 \small    07 June 1980\\
\hline
\small\bf SAKALAVA &
\small HANTASOA Marie Edvige& \small 02 November 1985\\
\small \bf  (MAINTIRANO)  &\small KOTOVAO Bernard &
 \small    06 October 1983\\
\hline
\small\bf BETSIMISARAKA &
\small RASOLONANDRASANA Voahirana & \small 24 September 1985\\
\small \bf  (MAHANORO)  &\small ANDRIANANDRASANA Maurice &
 \small    03 April 1979\\
\hline
\end{tabular}
\end{table}

\hfill\eject

\newpage


\begin{thebibliography}{000}

\bibitem[Adelaar, 1995a]{Adelaar:1995a}
A. Adelaar,
{\it Asian roots of the Malagasy; A linguistic perspective.}
In: Bijdragen tot de Taal-, Land- en Volkenkunde 151, no: 3, Leiden, 325-356 (1995).

\bibitem[Adelaar, 1995b]{Adelaar:1995b}
A. Adelaar,
{\it Borneo as a Cross-Roads for Comparative Austronesian Linguistics.}
In {\it The Austronesians in history.}
J. F. Bellwood, and D. Tryon editors, 75-95 (1995).
Australian National University, ANU E Press.

\bibitem[Adelaar, 2009]{Adelaar:2009}
A. Adelaar,
{\it  Loanwords in Malagasy.} 
In {\it Loanwords in the World's Languages: A Comparative Handbook,} 
M. Haspelmath and U. Tadmor editors, 717-746 (2009). Berlin: De Gruyter Mouton.

\bibitem[Bakker {\it et al}, 2009]{Bakker:2009}
D. Bakker, A. M\"{u}ller, V. Vellupillai, S. Wichmann, C. H. Brown, P. Brown, D. Egerov, 
R. Mailhammer, A. Grant and E. W. Holman, 
{\it Adding typology to lexicostatistics: a combined approach to language classification.} 
Linguistic Typology {\bf 13}, 167-179 (2009).

\bibitem[Barban\c{c}on {\it et al}, 2006]{Barbancon:2006}
F. Barban\c{c}on, T. Warnow, S. N. Evans, D. Ringe, and L. Nakhleh, 
{\it An experimental study comparing linguistic phylogenetic reconstruction methods.} 
Paper presented at the conference Languages and Genes, UC Santa Barbara, 
September 8-10, (2006), http://www.cs.rice.edu/~nakhleh/Papers/UCSB09.pdf.

\bibitem[Blanchard {\it et al}, 2010a]{Blanchard:2010a}
Ph. Blanchard, F. Petroni, M. Serva and D. Volchenkov,
{\it Geometric Representations of Language Taxonomies.} 
Computer Speech \& Language doi:10.1016/j.csl.2010.05.003 
(published on-line 21 May 2010).

\bibitem[Blanchard {\it et al}, 2010b]{Blanchard:2010b}
Ph. Blanchard, J.-R. Dawin, D. Volchenkov,
{\it Markov Chains or the Game of Structure and Chance: From Complex Networks, 
to Language Evolution, to Musical Compositions.}
European Physical Journal - Special Topics {\bf 184}, 1-82 (2010).

\bibitem[Blench, 2010]{Blench:2010}
R. M. Blench,
{\it The vocabularies of Vazimba and Beosi:
do they represent the languages of the pre-Austronesian 
populations of Madagascar?}
Preprint, Cambridge (2010)

\bibitem[Blench and Walsh, 2009]{Blench and Walsh:2009}
R. M. Blench and M. Walsh,
{\it Faunal names in Malagasy: their etymologies and implications for
the prehistory of the East African Coast.}
In {\it Eleventh International Conference on
Austronesian Linguistics}
(11 ICAL), Aussois, France, (2009).

\bibitem[Dahl, 1951]{Dahl:1951}
O. C. Dahl,
{\it Malgache et Maanjan: une comparaison linguistique}. 
Olso: Egede Instituttet (1951).

\bibitem[Dahl, 1991]{Dahl:1991}
O. C. Dahl,
{\it Migration from Kalimantan to Madagascar.}
Oslo: The Institute of Comparative Research in Human Culture, Norwegian University Press.

\bibitem[Dyen, 1953]{Dyen:1953}
I. Dyen,
{\it\ Review of Otto Dahl, Malgache et Maanjan.} 
Language {\bf 29.4}, 577-590 (1953).

\bibitem[Faubl\'{e}e, 1983]{Faublee:1983}
J. Faubl\'ee,  
{\it M\'emoire sp\'ecial du Centre d'\'etudes sur le monde arabe et du Centre d'\'etudes sur 
l'oc\'ean occidental, 21-30 (1983).} 
Paris: INALCO \& Conseil International de la language fran\c{c}aise.

\bibitem[Fouracre:2007]{Fouracre:2007}
P. Fouracre,
{\it The New Cambridge Medieval History},
Cambridge University Press (1995-2007).

\bibitem[Greenhill {\it et al}, 2009]{Greenhill:2009}
S. J. Greenhill, R. Blust, and R.D. Gray,
{\it The Austronesian Basic Vocabulary Database.}
http://language.psy.auckland.ac.nz/austronesian, (2003-2009).

\bibitem[Gudschinsky, 1956]{Gudschinsky:1956}
S. Gudschinsky, 
{\it The ABC's of lexicostatistics (glottochronology).}
Word {\bf 12}, 175-210 (1956).

\bibitem[Heggarty, 2006]{Heggarty:2006}
P. Heggarty,
{\it Interdisciplinary indiscipline? Can phylogenetic methods meaningfully be applied 
to language data and to dating language?} 
In {\it Phylogenetic Methods and the Prehistory of Languages},    
P. Forster  and  C. Renfrew editors,  p. 183, 
McDonald Institute for Archaeological Research, Cambridge (2006). 

\bibitem[Holman {\it et al}, 2008]{Holman:2008}
E. W. Holman, S. Wichmann, C. H. Brown, V. Velupillai, A. M\"{u}ller, and D. Bakker,
{\it Explorations in automated language comparison.}
Folia Linguistica  {\bf 42.2},  331-354 (2008).

\bibitem[Houtman, 1603]{Houtman:1603}
F. Houtman,  
{\it Spraeckende woord-boeck inde Maleysche ende Madagascarsche talen met vele 
Arabische ende Turcsche woorden}. Amsterdam: Jan Evertsz (1603).

\bibitem[Hurles {\it et al} 2005]{Hurles:2005}
M. E. Hurles, B. C. Sykes, M. A. Jobling and P. Forster,
{\it The dual origin of the Malagasy in Island Southeast Asia and East Africa: 
Evidence from maternal and paternal lineages.} 
American Journal of Human Genetics {\bf 76}, 894-901 (2005).

\bibitem[Jolliffe, 2002]{Jolliffe:2002}
I. T. Jolliffe, {\it Principal Component Analysis}.
Springer Series in Statistics {\bf XXIX}, 2nd ed. (2002), Springer, NY.

\bibitem[Levenshtein, 1966]{Levenshtein:1966}
V. I. Levenshtein, 
{\it Binary codes capable of correcting deletions, insertions and reversals.} 
Soviet Physics Doklady {\bf 10}, 707 (1966).

\bibitem[Mariano, 1613-14]{Mariano:1613-14}
L. Mariano, 
{\it  R\'elation du voyage de decouverte fait \`a l'\^ile Saint-Laurent 
dans les ann\'ees  1613-1614.}
In {\it Collection des ouvrages anciens concernant Madagascar}, 
ed. by A. and G. Grandidier, 2: 1-64 (1613-1614). Paris, Comit\'e de Madagascar.

\bibitem[Petroni and Serva, 2008]{Petroni:2008}
F. Petroni and M. Serva, 
{\it Languages distance and tree reconstruction.} 
Journal of Statistical Mechanics: theory and experiment, P08012 (2008).

\bibitem[Petroni and Serva, 2010]{Petroni:2010}
F. Petroni and M. Serva,
{\it Measures of lexical distance between languages.} 
Physica A {\bf 389}, 2280-2283 (2010).

\bibitem[Saitou and Nei, 1987]{Saitou:1987}
N. Saitou and M. Nei,
{\it The neighbor-joining method: a new method for reconstructing phylogenetic trees.}
Molecular Biology and Evolution {\bf 40}, 406-425 (1987).

\bibitem[Sapir, 1916]{Sapir:1916}
E. Sapir, 
{\it Time Perspective in Aboriginal American Culture, a Study in Method.} 
Geological Survey Memoir {\bf 90}: No. 13 (1916).
Anthropological Series. Ottawa: Government Printing Bureau.

\bibitem[Sch\"{o}lkopf {\it et al}, 1998]{Schoelkopf:1998}
B. Sch\"{o}lkopf, A.J. Smola and K.-R. M\"{u}ller,
{\it Nonlinear component analysis as a kernel eigenvalue problem.} 
Neural Computation {\bf  10}, 1299 (1998).

\bibitem[Serva and Petroni, 2008]{Serva:2008}
M. Serva and F. Petroni,
{\it  Indo-European languages tree by Levenshtein distance.}
EuroPhysics Letters {\bf 81}, 68005 (2008).

\bibitem[Serva and Petroni, 2011]{Serva:2011}
M. Serva and F. Petroni,
{\it Dialects of Malagasy.} (2011).
http://univaq.it/~serva/languages/languages.html.

\bibitem[Sokal and Michener, 1958]{Sokal:1958}
R. Sokal and C. D. Michener,
{\it A statistical method for evaluating systematic relationships.}
University of Kansas Science Bulletin {\bf 38}, 1409-1438 (1958).

\bibitem[Tuuk, 1864]{Tuuk:1864}
H. N. van der Tuuk,
{\it  Outlines of grammar of Malagasy language.}
Journal of the Royal Asiatic Society {\bf 8.2} (1864).

\bibitem[Vavilov, 1926]{Vavilov:1926}
N. I. Vavilov,
{\it Centers of origin of cultivated plants. }
Trudi po Prikl. Bot. Genet. Selek. 
[Bulletin of Applied Botany and Genetics] {\bf 16}, 139-248 (1926).

\bibitem[V\'{e}rin {\it et al}, 1969]{Verin:1969}
P. V\'{e}rin, C.P. Kottak and P. Gorlin,
{\it The glottochronology of Malagasy speech communities.} 
Oceanic Linguistics {\bf 8}, 26-83 (1969).

\bibitem[Wichmann {\it et al}, 2010a]{Wichmann:2010a}
S. Wichmann, A. M\"{u}ller and V. Velupillai,
{\it Homelands of the world's language families.} 
Diachronica {\bf 27.2}, 247-276 (2010).

\bibitem[Wichmann {\it et al} 2010b]{Wichmann:2010b}
S. Wichmann, E. W. Holman, D. Bakker and C.H. Brown,
{\it Evaluating linguistic distance measures.}
Physica A {\bf 389}, 3632-3639 (2010).

\bibitem[Wichmann {\it et al}, 2010c]{Wichmann:2010c}
S. Wichmann, A. M\"uller, V. Velupillai, C. H. Brown, E. W. Holman, 
P. Brown, S. Sauppe, O. Belyaev, M. Urban, Z. Molochieva, 
A.  Wett, D. Bakker, J-M List, D. Egorov, R. Mailhammer, 
D.Beck and H. Geyer,
{\it  The ASJP Database (version 13)}, (2010),
http://email.eva.mpg.de/~wichmann/languages.htm.


\end{thebibliography}
\end{document}